\documentclass{llncs}
\usepackage{graphicx}
\usepackage{comment}
\usepackage[caption=false]{subfig}
\usepackage{listings}
\usepackage{tikz}
\usetikzlibrary{trees}

\usepackage{color}

\lstdefinelanguage{XQuery}
{morekeywords={declare,function,for,let,where,return,as,if,then,else,at,in,is},
sensitive=true,
morecomment=[s]{(*}{*)},
morestring=[b][']
}

\lstdefinestyle{XML}{language=XML}
\lstdefinestyle{XQ}{language=XQuery,
basicstyle=\small,
keywordstyle=\bfseries,
emphstyle=\itshape}
\begin{document}

\title{A semantic approach for the requirement-driven discovery of web services in the Life Sciences}
\author{Mar\'ia P\'erez, Rafael Berlanga, Ismael Sanz}
\institute{Universitat Jaume I, Spain\\\email{\{mcatalan, berlanga, isanz\}@uji.es}}

\maketitle    

\begin{abstract}
Research in the Life Sciences depends on the integration of large, distributed and heterogeneous data sources and web services. The discovery of which of these resources are the most appropriate to solve a given task is a complex research question, since there is a large amount of plausible candidates and there is little, mostly unstructured, metadata to be able to decide among them. 
We contribute a semi-automatic approach, based on semantic techniques, to assist researchers in the discovery of the most appropriate web services to fulfill a set of given requirements. 
\end{abstract}

\section{Introduction}\label{sec:introduction}

Contemporary research in the Life Sciences depends on the sophisticated integration of large amounts of data obtained by in-house experiments 
 with reference databases available on the web
. This is followed by analysis workflows that rely on highly specific algorithms, often available as web services
. The amount of data produced and consumed by this process is prodigious; however, the sheer amount of available resources is a source of severe difficulties. 

Within this huge set of resources, one of the main problems to the user is finding the right web services for a given research
task. The landscape of Life Sciences web services is large
and complex: there are thousands of resources \cite{NARWS2010},  most of them available in public repositories, i.e. BioCatalogue \cite{BioCatalogue10}, 
but unfortunately only a few are described by adequate metadata, which is essentially textual in nature and this makes the discovery and the integration difficult. 
 In addition, there are many versions of different services that apparently provide the same broad functionality, but not enough metainformation is available to decide which of these services is the most appropriate for a precise task. 


Given this context, it is a pressing question how to help researchers to discover the best possible mapping between their requirements and the available tools. We present a semi-automatic approach to assist the researcher in web service discovery, looking for web services that are appropriate to fulfill the information requirements in the Life Sciences domain. The whole process is driven by well-captured requirements, in order to avoid the high costs associated with non-disciplined, non-reusable, ad-hoc development of integration applications. The matching between the requirements and web services is based on a semantic normalization of both the requirements and the web services metadata.  

\section{Approach overview}\label{sec:approach}
The overall approach we propose to assist the selection of web services based on users requirements consists of three main phases: (i) \emph{Requirements elicitation and specification}, (ii) \emph{Normalization}, and (iii) \emph{Web service selection}  phase. Here, we focus mainly on the \emph{normalization} and the \emph{web service selection} phases.

\begin{enumerate}

\item \textbf{Requirements elicitation and specification}.
The user's information requirements are the information that drives the discovery of web services
. Therefore, the  requirements are gathered and formally described  in the \emph{requirements model} using the \emph{i*} formalism \cite{Yu95PhD}. More details about this phase can be found on \cite{Perez09Requirements}. Figure \ref{fig:fragment_req} shows a fragment of the requirements model in which a subgoal and the tasks to achieve this subgoal are specified.

\begin{figure}[ht]
\begin{center}
\includegraphics[width=0.5\columnwidth]{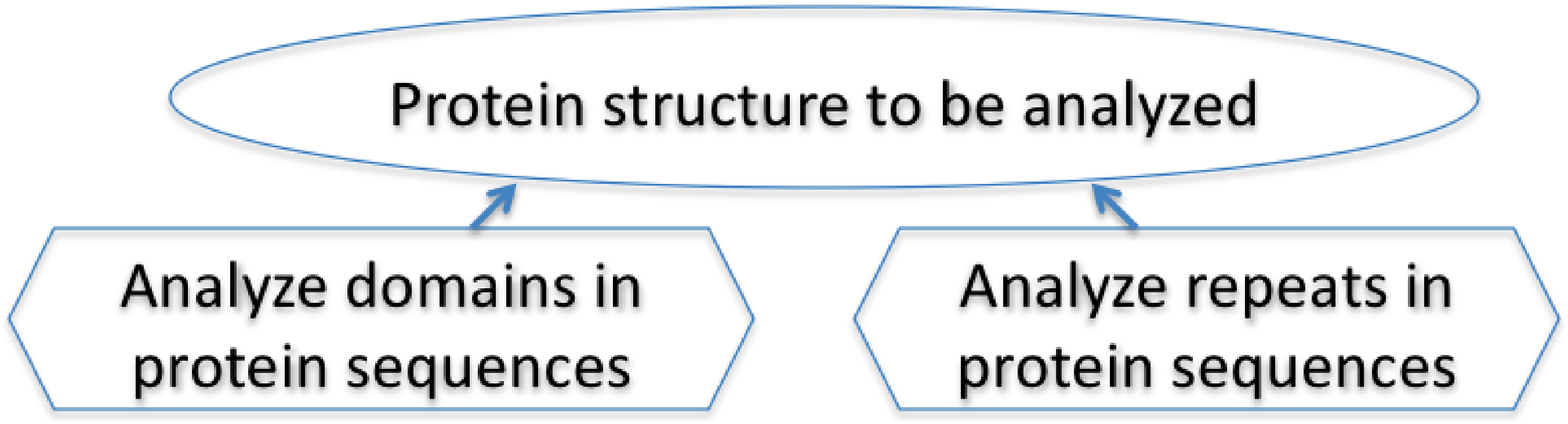}%
\caption{A fragment of the requirements model.}%
\label{fig:fragment_req}%
\end{center}
\end{figure} 

\item \textbf{Normalization}.
In the requirements model, task descriptions are expressed in natural language and, therefore, they must be normalized in order to be automatically processed. The normalization consists of a semantic annotation process in which the description of the tasks are processed and annotated with concepts from a reference ontology to allow the reconciliation of the user's requirements  with the web services. It is carried out in two phases: 

\begin{enumerate}
\item \textbf{Domain specific annotations}.
The purpose of this step is to identify the terms of the user-defined task related to the  Life Sciences domain. We have selected  a semantic annotator which is capable of using several ontologies to annotate biomedical texts \cite{Berlanga10Semantic}.  In our case, it annotates the task description with concepts from the UMLS and $^{my}$Grid ontologies. 

To determine the similarity between a text and a concept this annotator uses the following  information-theoretic function:
\[sim(C,T) = max_{S \in lex(C)}(ratio(S,T)) \]
\[ratio(S,T) = \frac{idf(cw(S,T)) - missing(S,T)}{idf(S)} \]
\[missing(S,T) = (idf(S) - idf(cw(S,T))) \]

The function $ratio(S,T)$ defines the ratio between the achieved information evidence for the text $T$, in our case the task description, and the information encoded in the lexicon form $S$. The function $missing(S,T)$ is the amount of information contained in $S$ that have not been covered with $T$.
$idf(S)$ measures the relevance of the terms in the string $S$, and  $cw(S,T)$ is the set of terms in common between the concept string $S$ and the text $T$. 
The function $idf$ is defined as follows:
\[idf(S)= - \sum_{w \in S} log ( P(w|UMLS) ) \] 
 
The relevance of word is measured by means of its estimated probability within the whole UMLS lexicon (i.e. $P(w|UMLS)$). After applying the semantic annotator, each task is represented with a semantic vector with the $tf*idf$ values of each concept. For example, the task ``Analyze domains in protein sequences'' is represented with the semantic vector \emph{\{'C1513868':8, 'D9000419':15\}} where C1513868 is retrieved by ``domains'' and D9000419 is retrieved by ``protein sequences''.

\item \textbf{Application specific annotations}.
The next step is to determine the  functionality the task is describing. For that purpose, we use the taxonomy of categories defined by BioCatalogue to classify the user-defined tasks. 
 The matching between the tasks and the categories  is made by using the ISub metric \cite{Stoilos05String} that performs a string matching algorithm.
For example, the task ``Analyze domains in protein sequences'' is annotated with the category ``Protein Sequence Analysis''  with a score of 0.5586.
\end{enumerate}

\item \textbf{Web services selection}.
 The discovery of the suitable web services for the user's requirements is based on the matching between the annotations of the tasks and the metadata of the web services. 
However,  most web service registries suffer from the lack of metadata. To address this problem, we have automatically annotated the description (or the documentation in case there is no description available), the categories  and the tags of  the 1729 web services registered in BioCatalogue. 
The annotations of each service are  stored as a vector that contains the $tf*idf$ of each concept. 

The discovery process is made by two independent searches: (i) search the web services that are annotated with the same category as the task, and (ii) search the web services that have concepts in common with the user-defined task. The results of both searches are combined and  scored based on the following linear combination:
\begin{center}
$ score=C$-$score*w_1 + S$-$score *w_2 $
\end{center}
where $w_1$ and $w_2$ are weights that depend on the relevance of each search and fulfill the condition $w_1+w_2=1$,  \emph{C-score} is the score calculated by the ISub metric, that is, the score of the string matching between the categories of the taxonomy and the user-defined task, and
\emph{S-score} is the cosine similarity between the vector of the task and the vector of the service. 

Table \ref{table:ranked_list_services} shows the highest ranked services for the task ``Analyze domains in protein sequences''.
\begin{table}
\begin{centering}
\begin{tabular}{|l|l|l|l|l|}
\hline 
Service & Shared annotations  & C-score & S-score & Score  \\
\hline
GlobPlot &  C1513868, D9000419 & 0 & 0.6934 & 0.5547 \\
Uniprot &   D9000419 & 0.5586 & 0.5427 & 0.5459 \\
GenesilicoProteinSilicoSOAP  &  C1513868, D9000419 & 0.5586 & 0.4725  & 0.4897 \\
Emboss tmap &   D9000419 & 0.5586 & 0.4443 & 0.4671 \\
ELMdb &  D9000419  & 0.5586 & 0.4379  & 0.4621 \\
\hline
\end{tabular} 
\caption{Ranked list of services for the task \emph{Analyze domains in protein sequences}.}
\label{table:ranked_list_services}
\end{centering}
\end{table}
\end{enumerate}
At the end, the user obtains a set of ranked lists of web services that are supposed to provide the functionality required by the tasks. 
In case the results are not those expected by the user, she can refine the process at the three phases of the guide.

\section{Conclusions and Future work}\label{sec:conclusions}
We have presented a semi-automatic approach that guides researchers in the Life Sciences to the discovery of web services that respond to their informational requirements
. 
Due to the importance of the semantic normalization, we have annotated the available information of the services registered in BioCatalogue by using a biomedical semantic annotator. In turn, the user requirements are also annotated with the same ontologies, thus allowing the application of a semantic search technique to find mappings between requirements and services.  
The result of the process is that the user is provided with a set of ranked lists of web services that are appropriate for her stated information needs. 


Some direct follow-ups of this work are the refinement of some details of the semantic techniques, the exploitation of other sources of metadata (bibliographical  information, referenced web pages), and the creation of a GUI to facilitate its application.

\section*{Acknowledgements}
This research has been supported by the Spanish Ministry of Education and Science (grant TIN2008-01825/TIN) and by Universitat Jaume~I -- Fundaci\'o Bancaixa (grant P11B2008-43). Mar\'ia P\'erez has been supported by Universitat Jaume~I predoctoral grant PREDOC/2007/41.

\bibliographystyle{plain}
\bibliography{swat4ls}
\end{document}